\documentclass[sigconf]{acmart}
\AtBeginDocument{%
  }

\usepackage{microtype}
\usepackage{graphicx}
\usepackage{adjustbox}
\usepackage{booktabs}

\usepackage{hyperref}

\usepackage{amsmath}
\usepackage{amssymb}
\usepackage{mathtools}
\usepackage{amsthm}

\usepackage[textsize=tiny]{todonotes}

\usepackage{xspace}

\usepackage[utf8]{inputenc} % allow utf-8 input
\usepackage[T1]{fontenc}    % use 8-bit T1 fonts
\usepackage{hyperref}       % hyperlinks
\usepackage{url}            % simple URL typesetting
\usepackage{booktabs}       % professional-quality tables
\usepackage{amsfonts}       % blackboard math symbols
\usepackage{nicefrac}       % compact symbols for 1/2, etc.
\usepackage{microtype}      % microtypography
\usepackage{xcolor}         % colors

\usepackage{microtype}
\usepackage{graphicx}
\usepackage{subfigure}
\usepackage{booktabs}
\usepackage{graphicx, array} % Required for inserting images
\usepackage{amssymb}% for professional tables
 \usepackage{threeparttable}
\usepackage{hyperref}
\usepackage{colortbl}
\usepackage{pifont}

\usepackage{amsmath}
\usepackage{amssymb}
\usepackage{mathtools}
\usepackage{amsthm}
\usepackage{algorithm,algorithmic}
\usepackage[capitalize,noabbrev]{cleveref}
\usepackage{multirow}
\usepackage{arydshln}

\usepackage[table]{xcolor}
\usepackage{subcaption}

\usepackage[most]{tcolorbox}
\tcbset{before skip=6pt, after skip=6pt}

\newtcolorbox{promptcard}[2][]{%
  enhanced,
  breakable,
  colback=white,
  colframe=black!30,
  boxrule=0.6pt,
  arc=2mm,
  left=1.2mm,right=1.2mm,top=1.0mm,bottom=1.0mm,
  fonttitle=\bfseries,
  title={#2},
  #1
}

\usepackage{gradient-text}
\newcommand{\concordia}{\gradientRGB{Concordia}{160,70,70}{140,170,210}\xspace}

\definecolor{NatureBlue}{RGB}{204,220,236}
\definecolor{RowGray}{gray}{0.92}
\definecolor{InkRed}{RGB}{210,120,120}
\newcommand{\mcc}[2]{\ensuremath{#1\!\pm\!#2}}

\definecolor{bgcolor}{rgb}{0.85,0.85,1}

\definecolor{mydarkgreen}{RGB}{39,130,67}
\definecolor{mydarkred}{RGB}{192,25,25}

\usepackage[table]{xcolor}
\colorlet{blue}{cyan!60}
\makeatletter
\newcommand{\algcolor}[2]{%
  \hskip-\ALG@thistlm\colorbox{#1}{\parbox{\dimexpr\linewidth-2\fboxsep}{\hskip\ALG@thistlm\relax #2}}%
}

\makeatother
\begin{document}

%%
%% The "title" command has an optional parameter,
%% allowing the author to define a "short title" to be used in page headers.
\title{\concordia: Self-Improving Synthetic Tables for Federated LLMs}

%%
%% The "author" command and its associated commands are used to define
%% the authors and their affiliations.
%% Of note is the shared affiliation of the first two authors, and the
%% "authornote" and "authornotemark" commands
%% used to denote shared contribution to the research.
% \author{Ben Trovato}
% \authornote{Both authors contributed equally to this research.}
% \email{trovato@corporation.com}
% \orcid{1234-5678-9012}

\author{Jimin Huang}
\authornote{Corresponding author.}
% \authornotemark[1]
\email{jimin.huang@postgrad.manchester.ac.uk}
\affiliation{%
  \institution{University of Manchester}
  \city{}
  \country{}
}

\affiliation{%
  \institution{The Fin AI}
  \city{}
  \country{}
}
\author{Duanyu Feng}
\affiliation{%
  \institution{The Fin AI}
  \city{}
  \country{}
}

\author{Nuo Chen}
\affiliation{%
  \institution{National University of Singapore}
  \city{}
  \country{}
}

\author{Xiaoyu Wang}
\affiliation{%
  \institution{New York University}
  \city{}
  \country{}
}

\author{Zhiqiang Zhang}
\affiliation{%
  \institution{The Fin AI}
  \city{}
  \country{}
}

\author{Xueqing Peng}
\affiliation{%
  \institution{The Fin AI}
  \city{}
  \country{}
}

\author{Mingquan Lin}
\affiliation{%
  \institution{University of Minnesota}
  \city{}
  \country{}
}
\author{Prayag Tiwari}
\affiliation{%
  \institution{Halmstad University}
  \city{}
  \country{}
}
\author{Guojun Xiong}
\affiliation{%
  \institution{Harvard University}
  \city{}
  \country{}
}
\author{Alejandro Lopez-Lira}
\affiliation{%
  \institution{University of Florida}
  \city{}
  \country{}
}
\author{Sophia Ananiadou}
\affiliation{%
  \institution{University of Manchester}
  \city{}
  \country{}
}

%%
%% By default, the full list of authors will be used in the page
%% headers. Often, this list is too long, and will overlap
%% other information printed in the page headers. This command allows
%% the author to define a more concise list
%% of authors' names for this purpose.
% \renewcommand{\shortauthors}{Trovato et al.}

%%
%% The abstract is a short summary of the work to be presented in the
%% article.
\begin{abstract}
Federated learning (FL) enables training large language models (LLMs) without sharing raw data, but adapting LLMs under strict data isolation and non-IID client distributions remains challenging in practice.
Synthetic data offers a natural privacy-preserving surrogate for local training, yet existing federated pipelines typically treat synthetic generation as static or loosely coupled with downstream optimization, leading to rapidly diminishing utility under heterogeneous clients.
We study federated adaptation of LLMs on tabular tasks where raw records and validation data cannot be shared, and local training must rely entirely on synthetic tables.
We propose \textsc{\concordia}, a tri-level optimization framework that aligns synthetic data generation with federated validation utility despite these constraints.
At the client level, models are adapted via parameter-efficient LoRA training on synthetic tables.
Clients additionally learn lightweight utility scorers from private validation feedback to reweight synthetic samples during local training.
At the outer level, each client refines its own synthetic table generator using group-relative policy optimization (GRPO), guided by an ensemble of heterogeneous scorers shared across clients, without aggregating generator parameters or exposing validation data.
Experiments on privacy-sensitive tabular benchmarks from finance and healthcare demonstrate that \textsc{\concordia} consistently improves federated performance, cross-client stability, and robustness to distribution shift compared to static and decoupled synthetic-data baselines.

\end{abstract}

\renewcommand{\shortauthors}{Huang et al.} 
\maketitle

% % \input{intro_nuorevised}
\section{Introduction}
Adapting large language models (LLMs) in privacy sensitive and regulated domains is difficult because supervision is local, heterogeneous, and non IID, while raw records cannot be pooled.~\cite{Yan2024SmallMA,wang2024harmonic,Nam2024OptimizedFG} We study tabular prediction in finance and healthcare under a high-assurance setting where \textbf{no tables may leave an institution} (real or synthetic) and \textbf{high-capacity learned parameters are non-exportable}.~\cite{Chua2024MindTP,wang2025rewardds,Stoisser2025QueryDT} This setting is most challenging under extreme class imbalance, e.g., insurance risk prediction with positives near $1\%$ or below, where minority signals are crucial yet easy to wash out.~\cite{Weng2024ClassIB,Akinseloyin2025WeaklySA} LLM-based learners can consume natural-language feature definitions and policy rules, which helps bridge schema and policy shifts across institutions, but only if they can be adapted under these isolation constraints.

Federated learning is a natural baseline when model updates can be shared, but under non IID clients and extreme imbalance its aggregated optimization tends to follow dominant clients and frequent labels, washing out rare outcomes.~\cite{lomurno2024FedKR, abacha2024syntheticdata, goetz2020federated, hu2022fedsynthgradientcompressionsynthetic, xin2022flgan} In the high-assurance setting we target, exporting gradients or LoRA deltas is also typically prohibited, so standard update-sharing FL is either ineffective or infeasible. A common workaround is to train locally on \emph{on-premise} synthetic tables, and recent pipelines use LLMs to generate context-rich tabular samples.~\cite{borisov2022language, solatorio2023realtabformer, schmidhuber2024llm} Yet most synthetic approaches are either fixed, or they optimize generation with objectives that are decoupled from the evolving downstream model and client-specific distributions across rounds. Under heterogeneity, such proxy improvements in synthetic data do not reliably translate into gains on each client’s private validation set.~\cite{Ravuri2019ClassificationAS} We call this \textbf{the validation aligned synthetic optimization problem in heterogeneous federated learning under restricted communication}: synthetic refinements must predict gains in client local validation utility when neither real/synthetic tables nor high-capacity model or generator parameters can be shared. The key difficulty is that under non-IID clients and extreme imbalance, validation utility is sparse, noisy, and non-stationary across rounds, so naïvely improving synthetic data with proxy objectives does not yield reliable utility gains.

To address it, we propose \concordia, a tri-level federated loop that turns private validation utility into an online signal for refining the on-premise synthetic training distribution under heterogeneity.~\cite{Ren2018LearningTR} The three levels are jointly necessary: \textbf{local LoRA~\cite{Hu2021LoRALA} adaptation} defines the evolving client model state, \textbf{utility modeling} distills private validation feedback into a capacity-limited scorer that reweights synthetic mini-batches toward utility-critical (often minority) cases, and \textbf{federated coordination} stabilizes this utility proxy across non-IID clients. Concretely, each client alternates multiple LoRA steps on locally generated synthetic batches with intermittent scorer refresh on its private validation set, so the proxy tracks what improves utility as the model and distribution drift over rounds. At the federated level, \emph{only} scorer parameters are exchanged: the server pools and redistributes diverse scorers to reduce client-specific bias and high-variance utility noise, yielding a stable relative preference signal that each client uses to update its on-premise generator online. Unlike fixed synthetic or decoupled pipelines, \concordia closes the missing loop by letting validation utility continuously reshape the synthetic distribution seen by the next round, while keeping records, synthetic tables, generators, and LoRA weights strictly on premise.

We evaluate \concordia on privacy-constrained federated tabular prediction across finance and medicine, including extreme imbalance (Travel\footnote{https://www.kaggle.com/datasets/mhdzahier/travel-insurance} positives near $0.01$) and out-of-site testing on VA Long Beach~\cite{Detrano1989InternationalAO}. Under our high-assurance constraint, naïve single-node training often collapses on minority-heavy settings. \concordia delivers its clearest gains exactly there: on Travel, MCC~\cite{Matthews1975ComparisonOT} rises from \textbf{7.07} at Round~1 to \textbf{21.78} at Round~2, moving from near-degenerate behavior to reliably useful minority-sensitive prediction, while remaining competitive on German~\cite{statloggerman_credit_data144} and LendingClub\footnote{https://www.kaggle.com/datasets/wordsforthewise/lending-club}. On the medical benchmark, it achieves non-trivial generalization to the unseen VA site. Case studies and embedding visualizations further show that, over rounds, high-utility synthetic samples concentrate around real positive neighborhoods and hard positives become better covered, providing a concrete mechanism consistent with the observed gains under non-IID imbalance.

\section{Related Work}
\subsection{LLM-based tabular learning and data synthesis.}
Classical tabular synthesis relies on GANs, VAEs, and diffusion models \cite{zhao2021ctab, xu2019modeling, kotelnikov2023tabddpm}, but often depends on rigid encodings and can struggle to preserve feature semantics under high-dimensional schemas or schema variation. Recent work uses LLMs to serialize tables and generate schema-conditioned samples with stronger semantic consistency \cite{borisov2022language, solatorio2023realtabformer, schmidhuber2024llm}, which is appealing when feature definitions and policy rules are naturally expressed in language. However, LLM-based synthesis can still leak memorized information \cite{yan2024protecting} and is typically optimized with proxy objectives that are decoupled from an evolving downstream learner, making synthetic improvements unreliable under distribution shift, non-IID clients, and extreme imbalance. Our work builds on LLM-based synthesis but focuses on \emph{validation-aligned} and \emph{online} refinement of on-premise generation toward downstream utility in heterogeneous, privacy-constrained settings.

\subsection{Federated adaptation and personalization under heterogeneity.}
Federated personalization addresses non-IID clients by learning client-specific models or partially shared parameters, reducing the bias of a single global model toward dominant clients.~\cite{Tan2021TowardsPF,Collins2021ExploitingSR} However, most approaches focus on how to aggregate or personalize model updates given local supervision, and implicitly assume that such updates can be communicated.~\cite{Yi2023FedLoRAMP,Babakniya2023SLoRAFP,Fan2023FATELLMAI,Cheng2025TowardFL,Ye2024OpenFedLLMTL,Ye2024FedLLMBenchRB,Guo2025CanFL,Wang2024OptimizingCD} In our high-assurance setting, high-capacity updates are non-exportable and labels are extremely imbalanced, so the limiting factor is not only aggregation but the availability of utility-relevant training signal as the learner evolves. \concordia targets this bottleneck by using private validation utility to continuously reshape on-premise synthetic training toward minority-critical utility under heterogeneity.

\subsection{Federated learning with synthetic data and feedback.}
Synthetic data has been used in federated learning to reduce raw-data exposure, lower communication cost, and improve training efficiency by generating data locally (often under differential privacy) and either sharing synthetic samples or using them as a surrogate for sensitive local training \cite{lomurno2024FedKR, abacha2024syntheticdata, goetz2020federated, hu2022fedsynthgradientcompressionsynthetic, xin2022flgan}. In parallel, feedback-driven generation in NLP improves task relevance via reward modeling and downstream signals \cite{wang2025rewardds, hou2025private, hou2024pre, wu2024lanfl, zou2025contrastive}. For privacy-sensitive tabular learning, however, existing approaches often face two limitations: they are less reliable under strongly heterogeneous clients with extreme class imbalance \cite{swanberg2025api}, and generation is commonly fixed or only loosely coupled to the evolving federated learner, so the most informative signal is not exploited in a stable loop. Our setting further restricts communication, where neither real/synthetic tables nor high-capacity model or generator parameters can be shared. \concordia closes this gap by learning lightweight utility scorers from private validation data on each client and using a redistributed scorer pool as the only exchanged signal to guide on-premise, decentralized generator refinement toward minority-critical utility under non-IID clients.

\begin{figure*}[tb]
    \centering
    \includegraphics[width=0.95\linewidth]{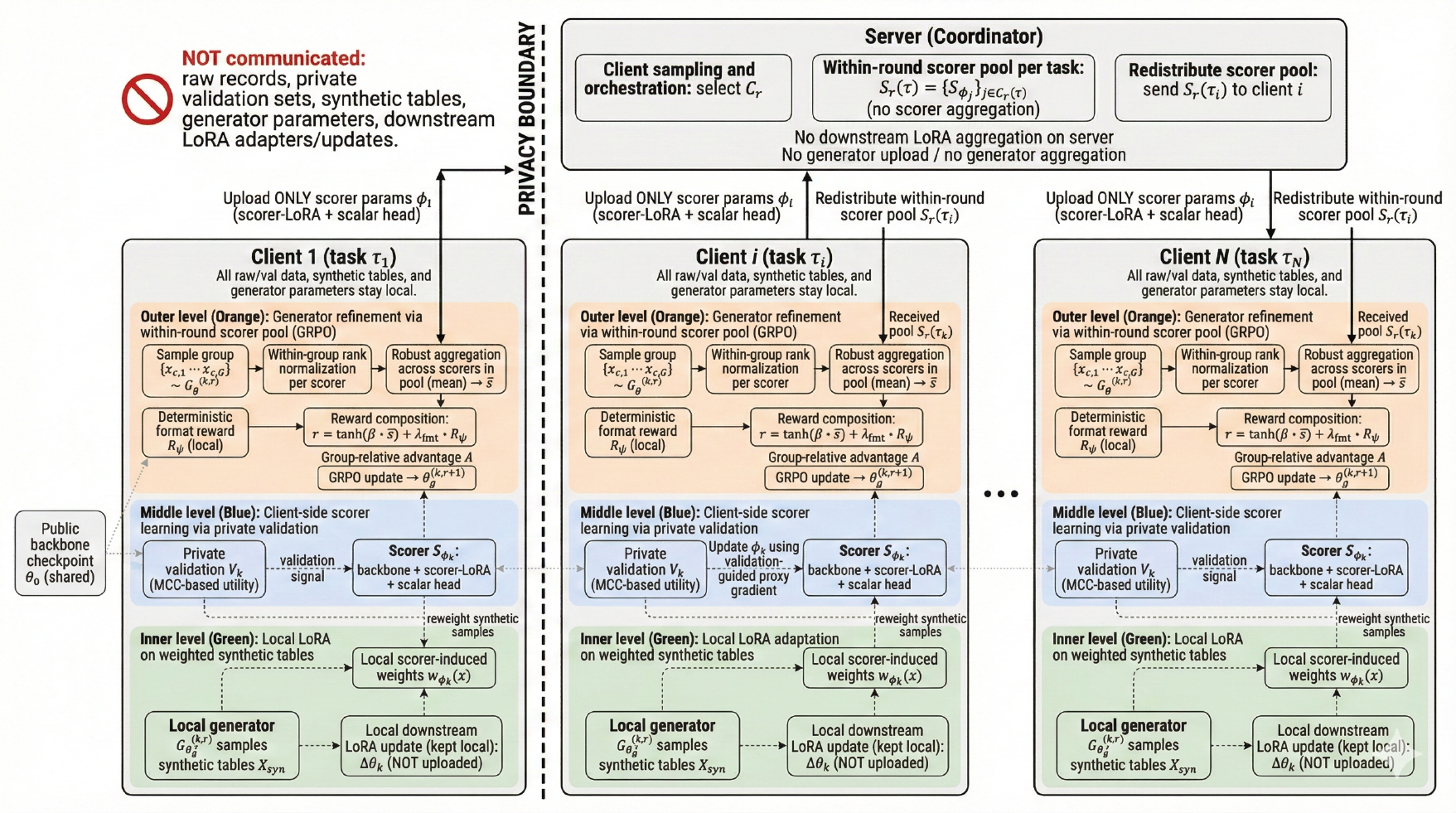}
    \caption{The Overall Framework for Concordia.}
    \label{fig:system}
\end{figure*}
\section{Problem Statement}
We study \emph{cross-silo federated learning} for adapting LLMs to tabular prediction under strict data-isolation rules in regulated deployments.
A system consists of $N$ clients indexed by $i\in[N]$ and a coordinating server that orchestrates multi-round communication.
Client $i$ holds a private dataset $\mathcal{D}_i=\{(x,y)\}$ and a private validation set $\mathcal{V}_i$, neither of which can be shared.

\paragraph{Motivating setting and heterogeneity.}
In privacy-sensitive domains such as finance and healthcare, the labels that matter most are often rare (e.g., default, fraud, adverse clinical outcomes).~\cite{Rieke2020TheFO}
Institutions differ in schema, feature semantics, decision policies, and label prevalence.~\cite{Collins2025FederatedLA}
We denote the client-specific positive rate by $p_i$ and allow it to vary widely across clients, including $p_i=0$.
When $p_i=0$ or $p_i$ is extremely small, naive local learning contains little minority signal and can collapse to majority-only behavior. Under non-IID clients, this failure is amplified and harms the sites where rare-event performance matters.
Accordingly, we emphasize long-tail utility after local adaptation, including worst-client behavior, rather than only average accuracy.

\paragraph{Tabular task and LLM interface.}
Each example consists of a mixed-type feature vector $x$ and a label $y\in\mathcal{Y}$.
To use an LLM as a predictor, client $i$ serializes $x$ into text using a schema-conditioned template $\mathcal{T}_i$, and the model predicts $y$ from the serialized input.
We consider both single-task federation (shared prediction target) and multi-task federation with task identifiers $\tau_i$.
Clients are in the same task if they share the same label space and prediction target; $\tau_i$ is treated as deployment metadata.
In our experiments, all medical sites share one task identifier, while each finance dataset is treated as a separate task identifier and trained as an independent federated stream to avoid ill-posed cross-task aggregation and negative transfer.

\paragraph{Communication and threat model (compliance-constrained FL).}
We consider a federated deployment where institutions are prohibited from exporting any data-dependent high-capacity artifacts.
At communication round $r$, the server broadcasts a shared backbone model $\theta^{(r)}$ to participating clients.
Client-to-server communication is restricted to a lightweight utility proxy (scorer parameters):
\begin{equation}
m_i^{(r)} \;=\; \phi_i^{(r)}.
\end{equation}
In particular, raw records, validation sets, synthetic tables, generator parameters, and any parameter-efficient adaptation updates (e.g., LoRA/adapter weights) are non-exportable and never transmitted.
Thus, unlike standard FL where the server aggregates uploaded model updates, our protocol uses the server only to pool and redistribute utility proxies across clients.
We consider adversaries that attempt to infer client-specific records or label properties from communicated artifacts; restricting communication to a small utility proxy reflects common audit and risk controls in regulated environments.
We do not claim a formal privacy guarantee and instead study a practical design that reduces direct exposure of records and validation data while enabling effective adaptation under heterogeneity and severe imbalance.

\paragraph{Synthetic training under data isolation.}
Since centralized supervision is prohibited, clients rely on on-premise synthetic tables for local adaptation.
Client $i$ maintains a client-local generator $G_{\psi_i}$ that produces labeled synthetic examples $(\tilde{x},\tilde{y})$ consistent with its schema and task.
We assume access to a client-local labeling interface $\mathcal{A}_i$ that assigns labels to synthetic samples using institution-specific resources (e.g., internal risk policies in finance, coding rules or guideline-based criteria in medicine).
This interface can be noisy, and we do not require any labeling mechanism to be shared across clients.
Generator parameters $\psi_i$ are never uploaded or aggregated.

\paragraph{Local adaptation and utility proxy.}
Given $\theta^{(r)}$ and the current generator $G_{\psi_i}^{(r)}$, client $i$ performs parameter-efficient local adaptation (e.g., LoRA) on synthetic mini-batches.
To prioritize utility-critical (often minority) synthetic cases, client $i$ maintains a scorer $S_{\phi_i^{(r)}}$ that induces nonnegative weights $w_{\phi_i^{(r)}}(\tilde{x},\tilde{y})$ and reweights the synthetic loss during adaptation.
For readability, we omit the round index $(r)$ when clear.
Let $\tilde{z}:=(\tilde{x},\tilde{y})$, $G_i := G_{\psi_i}$, and $w_i(\tilde{z}) := w_{\phi_i}(\tilde{z})$.
\begin{equation}
\Delta\theta_i^{*}
:= \arg\min_{\Delta\theta_i}\;
\mathbb{E}_{\tilde{z}\sim G_i}\!\left[
w_i(\tilde{z})\,
\ell\!\left(\tilde{z};\theta+\Delta\theta_i\right)
\right].
\end{equation}
where $\ell(\cdot)$ is a differentiable surrogate loss.
The scorer is trained and refreshed on-premise using the private validation set $\mathcal{V}_i$ under the current adapted state, so that it tracks which synthetic samples improve validation utility as the learner evolves.

\paragraph{Overall objective.}
Given a shared backbone $\theta$, our goal is to learn client-local generators $\{\psi_i\}_{i=1}^N$ such that, after local adaptation on synthetic tables, the resulting client models achieve high validation utility.
We write the objective as minimizing an imbalance-aware validation loss:
\begin{equation}
\min_{\{\psi_i\}}
\frac{1}{N}\sum_{i=1}^{N}
\mathcal{L}_{\mathrm{val}}^{(i)}
\!\left(
\theta + \Delta\theta_i^{*}(\theta,\psi_i,\phi_i^{*}(\theta,\psi_i));\ \mathcal{V}_i
\right),
\label{eq:overall_clean}
\end{equation}
where $\mathcal{L}_{\mathrm{val}}^{(i)}$ is aligned with imbalanced evaluation (e.g., $-\mathrm{MCC}$), and $\phi_i^{*}(\theta,\psi_i)$ denotes the scorer obtained by on-premise training using $\mathcal{V}_i$.

\paragraph{Tri-level structure and the role of federation.}
Optimizing Eq.~\eqref{eq:overall_clean} couples three components: (i) local adaptation on synthetic data (inner level), (ii) validation-guided learning of the utility proxy (middle level), and (iii) a federated outer loop that stabilizes utility feedback under non-IID clients.
Since $\mathcal{V}_i$ cannot be centralized and neither generators nor adapted model updates can be transmitted, generator refinement must be driven by an indirect, communicable proxy of validation utility.
Accordingly, clients upload only $\phi_i^{(r)}$. The server pools and redistributes these scorers to provide each client with a stabler relative preference signal for updating its on-premise generator across rounds.
This retains the defining structure of federated learning, while operating under stricter compliance constraints on what can be communicated.
\section{Proposed Method}

\begin{algorithm}[t]
\caption{\concordia (Route A): Federated learning with utility-proxy pooling and on-premise generator refinement}
\label{alg:concordia_routeA}
\small
\begin{algorithmic}[1]
\STATE \textbf{Input:} Clients $\{(\mathcal{D}_i,\mathcal{V}_i,\tau_i)\}_{i=1}^N$, task backbones $\{\theta_{\tau}\}$, client generators $\{\psi_i^{(0)}\}$, rounds $R$
\STATE \textbf{Hyperparameters:} local steps $T$, scorer refresh interval $m$, GRPO group size $G$, temperature $\beta$, format weight $\lambda_{\mathrm{fmt}}$

\FOR{round $r=0$ to $R-1$}
    \STATE Server samples clients $\mathcal{C}_r$ and partitions $\mathcal{C}_r(\tau)=\{i\in\mathcal{C}_r\mid \tau_i=\tau\}$
    \STATE Server broadcasts the approved backbone $\theta_{\tau_i}$ to each $i\in\mathcal{C}_r$ \COMMENT{fixed / externally updated}

    \STATE \textbf{Client local adaptation and scorer learning (on-premise, parallel for each $i\in\mathcal{C}_r$)}
    \STATE Initialize local adapter $\Delta\theta_i \leftarrow 0$ and scorer parameters $\phi_i$
    \FOR{$t=1$ to $T$}
        \STATE Sample synthetic mini-batch $X_{\mathrm{syn}} \sim G_{\psi_i^{(r)}}$
        \STATE Compute weights $w \leftarrow w_{\phi_i}(X_{\mathrm{syn}})$
        \STATE $\Delta\theta_i \leftarrow \Delta\theta_i - \eta \nabla_{\Delta\theta_i}\sum_{x\in X_{\mathrm{syn}}} w(x)\,\ell(x;\theta_{\tau_i}+\Delta\theta_i)$
        \IF{$t \bmod m = 0$}
            \STATE $q_i \leftarrow \mathcal{L}_{\mathrm{val}}^{(i)}(\theta_{\tau_i}+\Delta\theta_i;\mathcal{V}_i)$
            \STATE $\phi_i \leftarrow \phi_i - \eta_{\phi}\nabla_{\phi_i}\!\Big[q_i\cdot\sum_{x\in X_{\mathrm{syn}}} w_{\phi_i}(x)\,\ell(x;\theta_{\tau_i}+\Delta\theta_i)\Big]$
        \ENDIF
    \ENDFOR
    \STATE Client uploads \textbf{only} $\phi_i$ to server \COMMENT{no $\Delta\theta_i$, no data, no generators}

    \STATE \textbf{Server forms within-round scorer pools and redistributes}
    \FOR{each task $\tau$ with $\mathcal{C}_r(\tau)\neq\emptyset$}
        \STATE $\mathcal{S}_r(\tau)\leftarrow \{S_{\phi_j}\}_{j\in\mathcal{C}_r(\tau)}$
    \ENDFOR
    \STATE Server sends pool $\mathcal{S}_r(\tau_i)$ to each $i\in\mathcal{C}_r$

    \STATE \textbf{Client on-premise generator refinement (parallel for each $i\in\mathcal{C}_r$)}
    \STATE Let $\mathcal{S}_i \leftarrow \mathcal{S}_r(\tau_i)\cup\{S_{\phi_i}\}$ and sample conditions $\{c\}$
    \FOR{each condition $c$}
        \STATE Sample group $\{x_{c,1},\dots,x_{c,G}\}\sim G_{\psi_i^{(r)}}(\cdot\mid c)$
        \FOR{$j=1$ to $G$}
            \STATE $\bar{s}_{c,j}\leftarrow \mathrm{RobustAgg}_{S\in\mathcal{S}_i}\big(\mathrm{NormGroup}(\{S(x_{c,k})\}_{k=1}^G)_j\big)$
            \STATE $r_{c,j}\leftarrow \tanh(\beta\,\bar{s}_{c,j}) + \lambda_{\mathrm{fmt}}\,R_{\mathrm{fmt}}(c,x_{c,j})$
        \ENDFOR
        \STATE $A_{c,j}\leftarrow r_{c,j}-\frac{1}{G}\sum_{k=1}^{G}r_{c,k}$
    \ENDFOR
    \STATE $\psi_i^{(r+1)} \leftarrow \mathrm{GRPO}(\psi_i^{(r)};\{(x_{c,j},A_{c,j})\})$
\ENDFOR

\STATE \textbf{Return:} client generators $\{\psi_i^{(R)}\}$ and scorer pool dynamics $\{\phi_i\}$
\end{algorithmic}
\end{algorithm}

\subsection{Overview and Tri-level Decomposition}
\label{sec:overview}

\concordia is a compliance-constrained federated learning framework that couples client-local adaptation with on-premise synthetic generation through a tri-level feedback loop.
The key idea is to make \emph{private validation utility actionable} without exporting data, synthetic tables, generators, or high-capacity model updates: clients distill validation feedback into utility proxies (scorers), and the server pools and redistributes these proxies to stabilize utility signals under non-IID heterogeneity and extreme imbalance.
As a result, synthetic generation becomes an adaptive decision variable that is continuously shaped by utility, rather than a fixed preprocessing step.

\paragraph{Round protocol and communication boundary.}
At round $r$, the server samples clients $\mathcal{C}_r$ and groups them by task identifier $\tau$.
For each task, the server broadcasts an approved backbone snapshot $\theta_{\tau}$ (fixed or externally updated by the deployment).
Each client uploads \emph{only} scorer parameters $\phi_i$; raw records, validation sets, synthetic tables, generator parameters, and any parameter-efficient adaptation weights (e.g., LoRA/adapter deltas) never leave the client.
The server uses uploaded scorers solely as utility proxies and redistributes a within-round scorer pool to participating clients.

\paragraph{Tri-level decomposition.}
Each round consists of three coupled updates.
\textit{Inner level:} client $i$ performs on-premise parameter-efficient adaptation on synthetic mini-batches sampled from its local generator, producing a local adapted state used only for evaluation and learning signals.
\textit{Middle level:} using private validation feedback, the client updates a lightweight scorer $S_{\phi_i}$ so that it assigns higher weight to synthetic samples that improve imbalance-aware validation utility.
\textit{Outer (federated) level:} the server pools scorers within each task to form $\mathcal{S}_r(\tau)=\{S_{\phi_j}\}_{j\in\mathcal{C}_r(\tau)}$ and redistributes $\mathcal{S}_r(\tau_i)$ to each client; client $i$ then refines its on-premise generator for the next round using preference/reward signals aggregated from the pooled scorers.

\subsection{On-premise LoRA Adaptation on Weighted Synthetic Tables}
\label{sec:lora_local}

Client $i$ samples a synthetic mini-batch $X_{\mathrm{syn}} \sim G_{\theta_g^{(i,r)}}$ and performs parameter-efficient adaptation (e.g., LoRA) locally by minimizing
\begin{equation}
\min_{\Delta\theta_i}\ 
\mathbb{E}_{x \sim G_{\theta_g^{(i,r)}}}
\left[w_{\phi_i}(x)\,\ell\!\left(x;\theta_{\tau_i}+\Delta\theta_i\right)\right],
\label{eq:lora_local}
\end{equation}
where $\Delta\theta_i$ never leaves the client and is used only to define the current adapted state for validation feedback and scorer learning.

\paragraph{Default weighting rule.}
Weights are computed within each mini-batch as
\begin{equation}
w_{\phi_i}(X_{\mathrm{syn}})
=
\mathrm{Norm}\!\left(
\left\{
\mathrm{clip}\!\left(\frac{S_{\phi_i}(x)+1}{2},\,w_{\min},\,w_{\max}\right)
\right\}_{x\in X_{\mathrm{syn}}}
\right),
\label{eq:weight_default_batch}
\end{equation}
where $\mathrm{Norm}$ normalizes weights to sum to one within the mini-batch.

\paragraph{What is federated in our setting.}
Unlike standard FedLoRA~\cite{Yi2023FedLoRAMP}, our compliance constraint prohibits uploading any adapter/LoRA deltas.
Instead, the only federated signal is the scorer: after local adaptation and validation-guided scorer updates, client $i$ uploads $\phi_i$ for within-task pooling and redistribution (Section~\ref{sec:overview}), which provides a stabler preference signal for on-premise generator refinement under non-IID imbalance.

\subsection{Client-side Scorer Learning via Private Validation Feedback}
\label{sec:bilevel_scorer}

Client $i$ maintains an \emph{exportable evaluator} (scorer) $S_{\phi_i}$ that maps a synthetic table $x$ to a scalar score and induces the reweighting rule in Eq.~\eqref{eq:weight_default_batch}.
Different from standard federated optimization that uploads task-model updates, our compliance constraint permits exporting only this evaluator. We denote by $\phi_i$ the \emph{full} scorer parameters, including its LLM backbone and scorer head, and $\phi_i$ is the only learned artifact that may leave the silo.

\paragraph{Scorer architecture.}
Given a synthetic table $x$ (optionally linearized with task schema and condition $c$), the scorer encodes it with an LLM and obtains a representation $h_{\phi_i}(x)$.
A small head then predicts a scalar logit $\hat{s}_{\phi_i}(x)=\mathrm{Head}_{\phi_i}(h_{\phi_i}(x))$.
To keep scores comparable across clients and stable for additive composition with the format reward, we use a bounded output
\begin{equation}
S_{\phi_i}(x) \;=\; \tanh\!\left(\hat{s}_{\phi_i}(x)\right)\in[-1,1].
\label{eq:scorer_tanh}
\end{equation}

\paragraph{Private validation signal.}
We use an imbalance-aware validation loss $\mathcal{L}_{\mathrm{val}}^{(i)}=1-\mathrm{MCC}$ in all experiments.
Every $m$ local steps, client $i$ evaluates the current locally adapted state and computes
\begin{equation}
q_i \leftarrow \mathcal{L}_{\mathrm{val}}^{(i)}\!\left(\theta_{\tau_i}^{(r)}+\Delta\theta_i;\mathcal{V}_i\right).
\label{eq:val_signal}
\end{equation}
Importantly, the scorer never observes $\mathcal{V}_i$ directly. It receives only the scalar utility $q_i$ computed on-premise, which serves as a validation-aligned learning signal.

\paragraph{Validation-aligned scorer update (first-order surrogate).}
Using the current synthetic mini-batch $X_{\mathrm{syn}}$, we update the scorer by
\begin{equation}
\phi_i \leftarrow \phi_i - \eta_\phi \nabla_{\phi_i}
\left[
q_i \cdot
\sum_{x \in X_{\mathrm{syn}}}
w_{\phi_i}(x)\,
\ell\!\left(x;\theta_{\tau_i}^{(r)}+\Delta\theta_i\right)
\right].
\label{eq:scorer_update}
\end{equation}
This update can be viewed as a truncated hypergradient surrogate: it ties scorer learning to the validation utility of the \emph{current} inner-loop iterate while avoiding differentiation through the full inner-loop optimum.
To limit memorization and reduce leakage risk when exporting scorers, we constrain the scorer update budget per round.

\subsection{Client-local Generator Refinement via Pooled Scorer Feedback}
\label{sec:grpo}

\paragraph{Within-task scorer pooling.}
At round $r$, the server groups participating clients by task and forms a within-task pool
\begin{equation}
\mathcal{S}_r(\tau)=\{S_{\phi_j}\}_{j\in\mathcal{C}_r(\tau)} ,
\label{eq:scorer_pool_task}
\end{equation}
which is redistributed to all clients $i\in\mathcal{C}_r(\tau)$.
Pooling stabilizes utility feedback under non-IID heterogeneity and extreme imbalance: it reduces the variance of single-client preferences and yields a more reliable relative signal for refining the local generator.

\paragraph{Format reward and utility composition.}
In addition to scorer preferences, each client computes a deterministic format reward $R_{\psi}$ that checks schema validity (required fields, label constraints, numeric ranges, JSON structure).
For condition $c$ and candidate $x_{c,j}$,
\begin{equation}
r^{\mathrm{fmt}}_{c,j}=R_{\psi}(c,x_{c,j})\in[-1,1].
\label{eq:format_reward}
\end{equation}

\paragraph{Groupwise normalization and pooled preference.}
Client $i$ samples a condition $c$ and draws a candidate group $\{x_{c,1},\ldots,x_{c,G}\}\sim G_{\theta_g^{(i,r)}}(\cdot\mid c)$.
For each scorer $S\in\mathcal{S}_r(\tau_i)$ we compute a group-normalized score via a rank transform:
\begin{equation}
\tilde{s}^{S}_{c,j}=
\mathrm{NormGroup}\!\left(\{S(x_{c,1}),\ldots,S(x_{c,G})\}\right)_j .
\label{eq:group_norm_score}
\end{equation}
We then aggregate across scorers (mean by default) to obtain a pooled preference score:
\begin{equation}
\bar{s}_{c,j}=
\mathrm{RobustAgg}_{S\in\mathcal{S}_r(\tau_i)}\!\left(\tilde{s}^{S}_{c,j}\right).
\label{eq:ensemble_reward}
\end{equation}

\paragraph{GRPO with group-relative advantage.}
We combine pooled preference with the format reward:
\begin{equation}
r_{c,j}= \beta\,\bar{s}_{c,j}+\lambda_{\mathrm{fmt}}\,r^{\mathrm{fmt}}_{c,j},
\label{eq:total_reward}
\end{equation}
and compute the group-relative advantage
\begin{equation}
A_{c,j}=r_{c,j}-\frac{1}{G}\sum_{k=1}^{G}r_{c,k}.
\label{eq:advantage_total}
\end{equation}
Client $i$ clips and normalizes $\{A_{c,j}\}_{j=1}^G$ within each group and updates $\theta_g^{(i,r)}$ via GRPO to obtain $\theta_g^{(i,r+1)}$.

\begin{table*}[t]
\centering
\scriptsize
\setlength{\tabcolsep}{3.0pt}
\renewcommand{\arraystretch}{1.08}

\begin{minipage}[t]{0.49\textwidth}
\centering
\textbf{Medical.} MCC mean $\pm$ std. \par\smallskip
\begin{tabular}{@{}lcccc@{}}
\toprule
\textbf{Model} & \textbf{Cleveland} & \textbf{Hungarian} & \textbf{Switzerland} & \textbf{VA} \\
\midrule
\rowcolor{RowGray}\multicolumn{5}{c}{\textbf{Baseline (Single-node)}}\\
\midrule
\textbf{Qwen3-4B} &
\cellcolor{NatureBlue!80.12}\mcc{60.25}{14.76} &
\cellcolor{NatureBlue!61.78}\mcc{23.57}{19.71} &
\cellcolor{NatureBlue!56.16}\mcc{12.31}{9.18} &
\cellcolor{NatureBlue!63.75}\mcc{27.50}{10.57} \\
\textbf{Qwen3-4B-Instruct} &
\cellcolor{NatureBlue!66.14}\mcc{32.28}{17.36} &
\cellcolor{NatureBlue!63.33}\mcc{26.65}{14.76} &
\cellcolor{NatureBlue!50.00}\mcc{0.00}{0.00} &
\cellcolor{NatureBlue!49.95}\mcc{-0.10}{0.08} \\
\textbf{Cleveland} &
\cellcolor{NatureBlue!50}\mcc{0.00}{0.00} &
\cellcolor{NatureBlue!50}\mcc{0.00}{0.00} &
\cellcolor{NatureBlue!50}\mcc{0.00}{0.00} &
\cellcolor{NatureBlue!50}\mcc{0.00}{0.00} \\
\textbf{Hungarian} &
\cellcolor{NatureBlue!50}\mcc{0.00}{0.00} &
\cellcolor{NatureBlue!50}\mcc{0.00}{0.00} &
\cellcolor{NatureBlue!50}\mcc{0.00}{0.00} &
\cellcolor{NatureBlue!50}\mcc{0.00}{0.00} \\
\textbf{Switzerland} &
\cellcolor{NatureBlue!50}\mcc{0.00}{0.00} &
\cellcolor{NatureBlue!50}\mcc{0.00}{0.00} &
\cellcolor{NatureBlue!50}\mcc{0.00}{0.00} &
\cellcolor{NatureBlue!50}\mcc{0.00}{0.00} \\
\midrule
\rowcolor{RowGray}\multicolumn{5}{c}{\textbf{Qwen3-4B (Round 1)}}\\
\midrule
\textbf{Cleveland} &
\cellcolor{NatureBlue!87.34}\mcc{74.69}{11.70} &
\cellcolor{NatureBlue!65.43}\mcc{30.86}{18.78} &
\cellcolor{NatureBlue!38.32}\mcc{-23.35}{12.53} &
\cellcolor{NatureBlue!66.81}\mcc{33.61}{25.02} \\
\textbf{Hungarian} &
\cellcolor{NatureBlue!73.14}\mcc{46.28}{16.52} &
\cellcolor{NatureBlue!70.00}\mcc{40.00}{18.03} &
\cellcolor{NatureBlue!48.36}\mcc{-3.29}{27.31} &
\cellcolor{NatureBlue!86.38}\mcc{72.76}{21.15} \\
\textbf{Switzerland} &
\cellcolor{NatureBlue!50.00}\mcc{0.00}{0.00} &
\cellcolor{NatureBlue!45.11}\mcc{-9.78}{18.08} &
\cellcolor{NatureBlue!55.06}\mcc{10.11}{28.21} &
\cellcolor{NatureBlue!71.00}\mcc{42.01}{14.63} \\
% \textbf{Merged} &
% \cellcolor{NatureBlue!86.75}\mcc{73.50}{12.54} &
% \cellcolor{NatureBlue!67.12}\mcc{34.24}{18.17} &
% \cellcolor{NatureBlue!58.88}\mcc{17.77}{29.97} &
% \cellcolor{NatureBlue!70.21}\mcc{40.42}{26.65} \\
\midrule
\rowcolor{RowGray}\multicolumn{5}{c}{\textbf{Qwen3-4B (Round 2)}}\\
\midrule
\textbf{Cleveland} &
\cellcolor{NatureBlue!91.15}\mcc{82.29}{8.97} &
\cellcolor{NatureBlue!50.00}\mcc{0.00}{0.00} &
\cellcolor{NatureBlue!50.00}\mcc{0.00}{0.00} &
\cellcolor{NatureBlue!57.00}\mcc{14.00}{7.52} \\
\textbf{Hungarian} &
\cellcolor{NatureBlue!83.61}\mcc{67.22}{13.30} &
\cellcolor{NatureBlue!73.15}\mcc{46.29}{17.72} &
\cellcolor{NatureBlue!63.62}\mcc{27.25}{32.79} &
\cellcolor{NatureBlue!67.15}\mcc{34.30}{12.49} \\
\textbf{Switzerland} &
\cellcolor{NatureBlue!69.26}\mcc{38.52}{13.11} &
\cellcolor{NatureBlue!50.00}\mcc{0.00}{18.14} &
\cellcolor{NatureBlue!83.85}\mcc{67.70}{40.37} &
\cellcolor{NatureBlue!45.18}\mcc{-9.64}{6.74} \\
\bottomrule
\end{tabular}
\end{minipage}\hfill%
\begin{minipage}[t]{0.49\textwidth}
\centering
\textbf{Finance.} MCC mean $\pm$ std. \par\smallskip
\begin{tabular}{@{}lccc@{}}
\toprule
\textbf{Model} & \textbf{German} & \textbf{Lendingclub} & \textbf{Travel} \\
\midrule
\rowcolor{RowGray}\multicolumn{4}{c}{\textbf{Baseline (Single-node)}}\\
\midrule
\textbf{Qwen3-4B} &
\cellcolor{NatureBlue!53.97}\mcc{7.93}{5.48} &
\cellcolor{NatureBlue!49.53}\mcc{-0.94}{0.59} &
\cellcolor{NatureBlue!50.00}\mcc{0.00}{0.00} \\
\textbf{Qwen3-4B-Instruct} &
\cellcolor{NatureBlue!52.63}\mcc{5.17}{6.87} &
\cellcolor{NatureBlue!54.64}\mcc{9.27}{8.13} &
\cellcolor{NatureBlue!48.58}\mcc{-2.85}{2.86} \\
\textbf{German} &
\cellcolor{NatureBlue!50}\mcc{0.00}{0.00} &
\cellcolor{NatureBlue!50}\mcc{0.00}{0.00} &
\cellcolor{NatureBlue!50}\mcc{0.00}{0.00} \\
\textbf{Lendingclub} &
\cellcolor{NatureBlue!50}\mcc{0.00}{0.00} &
\cellcolor{NatureBlue!50}\mcc{0.00}{0.00} &
\cellcolor{NatureBlue!50}\mcc{0.00}{0.00} \\
\textbf{Travel} &
\cellcolor{NatureBlue!50}\mcc{0.00}{0.00} &
\cellcolor{NatureBlue!50}\mcc{0.00}{0.00} &
\cellcolor{NatureBlue!50}\mcc{0.00}{0.00} \\
\midrule
\rowcolor{RowGray}\multicolumn{4}{c}{\textbf{Qwen3-4B (Round 1)}}\\
\midrule
\textbf{German} &
\cellcolor{NatureBlue!65.22}\mcc{30.44}{7.06} &
\cellcolor{NatureBlue!50.00}\mcc{0.00}{0.00} &
\cellcolor{NatureBlue!50.00}\mcc{0.00}{0.00} \\
\textbf{Lendingclub} &
\cellcolor{NatureBlue!50.00}\mcc{0.00}{0.00} &
\cellcolor{NatureBlue!89.92}\mcc{79.83}{1.49} &
\cellcolor{NatureBlue!50.00}\mcc{0.00}{0.00} \\
\textbf{Travel} &
\cellcolor{NatureBlue!50.00}\mcc{0.00}{0.00} &
\cellcolor{NatureBlue!50.00}\mcc{0.00}{0.00} &
\cellcolor{NatureBlue!53.53}\mcc{7.07}{1.85} \\
\midrule
\rowcolor{RowGray}\multicolumn{4}{c}{\textbf{Qwen3-4B (Round 2)}}\\
\midrule
\textbf{German} &
\cellcolor{NatureBlue!66.46}\mcc{32.93}{6.00} &
\cellcolor{NatureBlue!50.00}\mcc{0.00}{0.00} &
\cellcolor{NatureBlue!50.00}\mcc{0.00}{0.00} \\
\textbf{Lendingclub} &
\cellcolor{NatureBlue!50.00}\mcc{0.00}{0.00} &
\cellcolor{NatureBlue!95.11}\mcc{90.22}{1.05} &
\cellcolor{NatureBlue!50.00}\mcc{0.00}{0.00} \\
\textbf{Travel} &
\cellcolor{NatureBlue!50.00}\mcc{0.00}{0.00} &
\cellcolor{NatureBlue!50.00}\mcc{0.00}{0.00} &
\cellcolor{NatureBlue!60.89}\mcc{21.78}{5.84} \\
\bottomrule
\end{tabular}
\end{minipage}

\caption{Main results split by domain.}
\label{tab:main_results_split}
\end{table*}

% \textbf{Metrics}: F1
\section{Experiments}
\label{sec:experiments}

We evaluate \concordia under a compliance-constrained federated setting where neither training records nor validation sets can be centralized.
Our benchmarks stress three deployment frictions that dominate real rollouts: (i) extreme label imbalance (Travel Insurance, $p\approx 0.01$), (ii) cross-institution shift with an unseen medical site (VA Long Beach), and (iii) heterogeneous objectives across financial tasks where naive parameter merging induces negative transfer.

\subsection{Experimental Settings}
\label{sec:exp_settings}

\paragraph{Datasets and clients.}
Table~\ref{tab:datasets} summarizes the datasets and splits.
We evaluate two regimes.

First, a multi site single task regime in healthcare.
Each medical site is one client under the same prediction task.
Cleveland, Hungary, and Switzerland~\cite{Detrano1989InternationalAO} participate in federated training, while VA Long Beach is held out and used only for evaluation as an unseen site.

Second, a multi task regime in finance.
German Credit, Lending Club, and Travel Insurance are treated as three separate tasks.
Each dataset is treated as one client, so there is no within task client aggregation in finance.
This regime isolates cross task negative transfer and tests whether our evaluator sharing can still improve generator refinement when model merging is not applicable.

\begin{table}[t]
\centering
\small
\begin{adjustbox}{max width=\linewidth}
\begin{tabular}{l c c c}
\toprule
\textbf{Dataset} &
\textbf{Split (Tr/Val/Te)} &
\textbf{Features} &
\textbf{Pos. Ratio} \\
\midrule
\multicolumn{4}{l}{\textbf{Finance}} \\
\midrule
German Credit    & 700 / 100 / 200         & 21 & 0.29 \\
Lending Club     & 9,417 / 1,345 / 2,691   & 21 & 0.20 \\
Travel Insurance & 8,865 / 1,266 / 2,534   & 9  & 0.01 \\
\midrule
\multicolumn{4}{l}{\textbf{Healthcare}} \\
\midrule
Cleveland        & 231 / 41 / 31 & 13 & 0.45 \\
Hungary          & 224 / 40 / 30 & 13 & 0.35 \\
Switzerland      & 93 / 17 / 13  & 13 & 0.07 \\
VA Long Beach    & 153 / 27 / 20 & 13 & 0.28 \\
\bottomrule
\end{tabular}
\end{adjustbox}
\caption{Summary of tabular datasets used in our experiments.}
\label{tab:datasets}
\end{table}

\textbf{Finance.}
We use three tabular datasets for credit and risk assessment.
German Credit~\cite{misc_statlog_(german_credit_data)_144}.
Lending Club\footnote{https://www.kaggle.com/datasets/wordsforthewise/lending-club}.
Travel Insurance\footnote{https://www.kaggle.com/datasets/mhdzahier/travel-insurance}.

\textbf{Healthcare.}
We use the Heart Disease dataset~\cite{Detrano1989InternationalAO} comprising four medical sites.
Each site is treated as one client.

\paragraph{Federated protocol.}
We simulate $R=2$ rounds of coordination under a strict export policy: raw records, validation sets, synthetic tables, generators, and any task-model updates are non-exportable.
At each round, the server broadcasts the task backbone to participating clients.
Each client performs \emph{on-premise} LoRA adaptation on synthetic tables generated by its client-local generator, and updates an \emph{exportable scorer} using private validation feedback.
The only client-to-server message is the scorer parameters; the server forms a within-round scorer pool from participating clients and redistributes it, after which each client refines its generator locally via GRPO using rank-normalized pooled preferences.
Accordingly, our experimental comparisons focus on protocol-compatible settings that differ only in whether pooled scorer feedback is available, rather than on update-aggregation baselines that require transmitting model updates.

In cross-silo deployments in healthcare and finance, each communication round typically incurs non-trivial operational overhead (governance approval, secure execution, and coordination across institutions), making long-horizon iterative FL hard to sustain in practice.
We therefore focus on one- and two-round coordination as a realistic regime and evaluate whether pooled evaluator feedback can deliver measurable gains under extremely limited rounds.

\paragraph{Models and initialization.}
All clients initialize their generators from the same pretrained instruction following language model, followed by public instruction tuning to improve schema adherence.
After initialization, generators are maintained and updated independently on each client.
The downstream model uses the same backbone capacity across all methods and is adapted via the same LoRA configuration.
Unless stated otherwise, all methods use the same synthetic sampling budget per round and comparable local compute.

\paragraph{Evaluation and early-round stability.}
We report Matthews Correlation Coefficient (MCC)~\cite{Matthews1975ComparisonOT}, which is robust to class imbalance, as mean$\pm$std over multiple seeds.
Because strict deployments often limit the number of communication rounds, we focus on \emph{early-round utility}: the worst-round MCC over the first two rounds and the round-to-round variance within each run, which directly reflect reliability under constrained coordination budgets.

\paragraph{Baselines.}
We report only protocol-compatible settings that appear in the main results.
\textbf{Single-node} evaluates the pretrained backbone (\textbf{Qwen3-4B}~\cite{qwen3technicalreport}) and its instruction-tuned variant (\textbf{Qwen3-4B-Instruct}~\cite{qwen3technicalreport}) without any cross-client interaction.
\textbf{Local-only} (rows named by the corresponding dataset/site, e.g., \textbf{Cleveland} or \textbf{German}) performs fully on-premise adaptation and generator refinement without receiving pooled scorers from other clients.
\textbf{Round 1/2} report the same protocol after one and two rounds of scorer pooling and redistribution.
We do not include update-aggregation baselines here because they require exporting task-model updates, which violates the export policy defining our setting.

\subsection{System-level Performance under One-/Two-shot Cross-silo Coordination}
\label{sec:main_results}

We study a one-/two-shot coordination regime that is common in cross-silo healthcare and finance, where each round is expensive to execute and audit.

\paragraph{Few-shot coordination helps most when the label signal is a long tail.}
The clearest effect appears under extreme imbalance.
On Travel Insurance ($p\approx0.01$), the single-node backbone collapses to $0.00\pm0.00$ MCC, but one shot already recovers a non-trivial signal ($7.07\pm1.85$) and the second shot amplifies it to $21.78\pm5.84$.
The shape of this improvement matters: the second shot contributes the majority of the gain, consistent with a mechanism where the first shot establishes a coarse decision boundary while the second benefits from a refined synthetic distribution that better exposes minority-supporting cases.

\paragraph{Improvements are not explained by model merging: finance isolates the effect of coordination.}
Even without any within-task aggregation (each finance task has a single client), two-shot coordination remains strongly effective.
German improves from $7.93\pm5.48$ to $32.93\pm6.00$, and Lending Club improves from $-0.94\pm0.59$ to $90.22\pm1.05$.
Moreover, high performance stays largely on-task: the Lending Club-adapted model is strong on Lending Club while remaining near $0$ on other tasks, reinforcing the decision to avoid naive cross-task merging under heterogeneous objectives.
Together, these patterns indicate that the observed gains are driven by the coordination signal used to refine synthetic training—rather than by aggregating task-model updates across clients or tasks.

\paragraph{Cross-site shift yields asymmetric transfer: a mechanism clue, not noise.}
In healthcare, within-site utility improves substantially on participating clients (e.g., Cleveland $60.25\pm14.76 \rightarrow 82.29\pm8.97$; Hungarian $23.57\pm19.71 \rightarrow 46.29\pm17.72$).
For the held-out VA site, the single-node baseline is $27.50\pm10.57$, yet transfer behavior is sharply asymmetric: the best one-shot transfer reaches $72.76\pm21.15$, while some two-shot transfer can regress (e.g., $34.30\pm12.49$ in the best Round-2 transfer setting, and negative transfer in others).
This heterogeneity is informative. It suggests that under distribution shift and limited rounds, synthetic refinement can trade off in-domain fitting and out-of-site robustness, and that understanding \emph{which} refinements move samples toward the unseen-site decision boundary is central.

\paragraph{Reliability signals are visible even in a two-shot budget.}
While we do not claim long-horizon convergence from two rounds, several tasks show reduced seed sensitivity alongside higher mean utility (e.g., Cleveland std drops from $14.76$ to $8.97$; Lending Club from $1.49$ to $1.05$ from Round 1 to Round 2).
Conversely, variance can increase when the minority signal is first recovered (e.g., Travel), which is expected when performance becomes sensitive to rare-event coverage.
These observations motivate a mechanistic analysis focused on how synthetic samples move between rounds.

\begin{table}[t]
\centering
\scriptsize
\setlength{\tabcolsep}{2.1pt}
\renewcommand{\arraystretch}{1.05}
\begin{adjustbox}{max width=\linewidth}
\begin{tabular}{lccccccccccc}
\toprule
\textbf{Cand.} & Age & Sex & RestECG & Angina & OldPeak & ST-slope & Vessels & Synth. label
& $\hat{s}_{\phi1}$ & $\hat{s}_{\phi2}$ & $\hat{s}_{\phi3}$ \\
\midrule
\#1 & 65 & M & LVH & yes & 2.10 & down & 0 & heartdisease
& 2.3281 & 4.8125 & 5.0000 \\

\#2 & 55 & M & normal & no & 0.00 & up & 1 & healthy
& \textbf{2.9844} & 4.1250 & 4.0938 \\
\bottomrule
\end{tabular}
\end{adjustbox}
\caption{\textbf{Clinical paired case with aligned scorer pool (Cleveland, $G{=}2$).}}
\label{tab:clinical_pair_pool_aligned}
\end{table}

\begin{table}[t]
\centering
\scriptsize
\setlength{\tabcolsep}{2.1pt}
\renewcommand{\arraystretch}{1.05}
\begin{adjustbox}{max width=\linewidth}
\begin{tabular}{lccccccccccc}
\toprule
\textbf{Cand.} & Agency & Product & Dur. & Dest. & Sales & Comm. & Age & Claim
& $\hat{s}_{\phi1}$ & $\hat{s}_{\phi2}$ & $\hat{s}_{\phi3}$ \\
\midrule
\#1 & c2b & silver\_plan & 60 & thailand & 45.00 & 13.50 & 35 & no
& 0.9297 & -5.1563 & -3.7500 \\

\#2 & ssi & silver\_plan & 60 & india & 25.50 & 7.65 & 55 & no
& 0.9297 & -5.1563 & \textbf{-3.0625} \\
\bottomrule
\end{tabular}
\end{adjustbox}
\caption{\textbf{Finance paired case with mixed-task scorer pool (Travel, $G{=}2$).}}
\label{tab:finance_pair_pool}
\end{table}

\subsection{Cross-domain Generator Audit: Reward-aligned Synthesis in Finance and Healthcare}
\label{sec:cross_domain_audit}

To make the main results less of a black box, we audit the generator itself and ask a simple question:
\emph{what does reward-aligned refinement change in the synthetic tables that clients actually train on?}

\paragraph{Controlled paired generation protocol.}
For each dataset, we reuse the same few-shot prompt and identical decoding hyperparameters (temperature/top-$p$/max length).
We then sample paired synthetic tables from the generator snapshots used in each round.
Each table is schema-valid by construction and is labeled locally via the client labeling interface.
To interpret refinement, we inspect the same reward signal used for generator updates: an on-task scorer trained from the client’s private validation feedback.

\paragraph{Healthcare (Cleveland): sharpening multi-feature risk conjunctions under a fixed schema.}
For Heart Disease, the audit shows that refinement organizes synthetic samples into clinically interpretable archetypes rather than single-feature heuristics.
A high-risk profile jointly activates exercise-induced angina, elevated \texttt{OldPeak}, and adverse ST-segment slope patterns, and is labeled as \texttt{heartdisease}.
Conversely, a low-risk profile consistently exhibits \texttt{OldPeak}\,$\approx 0$, favorable slope, and normal ECG patterns, and is labeled as \texttt{healthy}.
The key point is that the label is supported by a \emph{conjunction} of features, not one attribute in isolation.
This is exactly the kind of structure that improves robustness under site shift: it captures transferable decision patterns that remain meaningful even when marginal feature distributions change, as in the unseen VA evaluation.

\paragraph{Finance (Travel): escaping majority-mode copying by preserving functional dependencies.}
Travel Insurance is an extreme long-tail setting ($p\approx 0.01$) and the few-shot context is dominated by \texttt{Claim\_Status=no}.
Under such skew, an unrefined generator can improve variety while still being utility-irrelevant, e.g., superficial changes that do not create new decision signal.
In contrast, the refined generator preserves the dominant mode (\texttt{airlines/online} with \texttt{silver\_plan}) \emph{while strengthening feature-level regularities that are meaningful for learning}.
In our paired examples, the generator consistently maintains a stable commission-to-sales relationship (\texttt{Commission}$=0.3\times$\texttt{Sales}) while varying destinations, duration, and age.
This is a concrete sign of reward alignment: refinement does not merely diversify strings or enforce formatting, but encourages internally consistent feature dependencies, which prevents the downstream learner from being dominated by synthetically frequent yet semantically noisy perturbations.

\paragraph{Takeaway.}
Across domains, refinement does not simply make outputs more formatted.
It reshapes the \emph{support} of synthetic training data in a reward-aligned way:
in finance, it preserves majority-mode plausibility while reinforcing coherent feature dependencies;
in healthcare, it sharpens semantically grounded risk conjunctions.

\begin{figure}[htb]
    \centering
    \includegraphics[width=\linewidth]{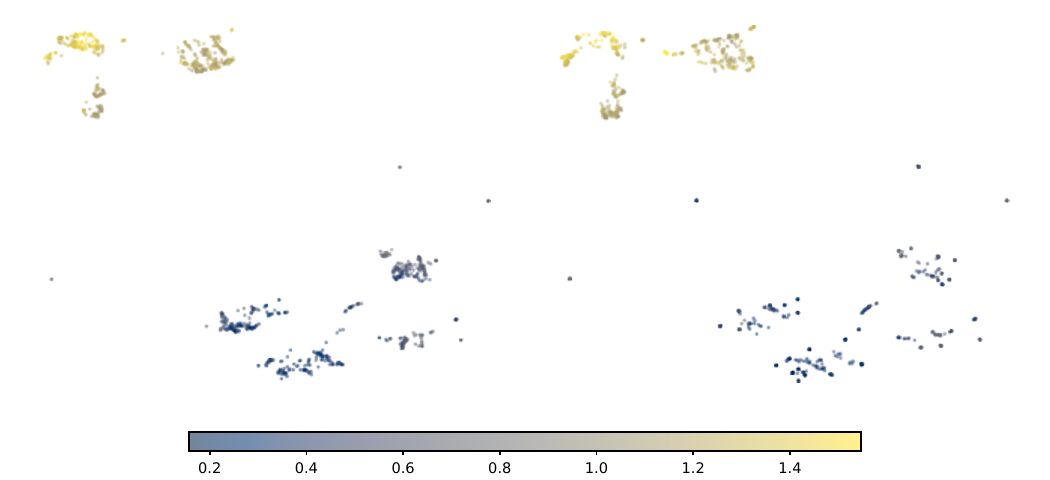}
    \caption{\textbf{Travel: reward-aligned refinement expands utility-relevant support.}}
\label{fig:travel_reward_vis}
\end{figure}

\subsection{Mechanistic Evidence on Travel: Reward-aligned Coverage Expansion}
\label{sec:mech_travel}

Travel Insurance is the most imbalanced task in our study ($p\approx0.01$) and exhibits the characteristic few-shot shape in the main results: MCC improves from $0.00\pm0.00$ (single-node) to $7.07\pm1.85$ after one shot, with a substantially larger gain to $21.78\pm5.84$ after the second shot.
We use the reward-colored embedding visualization in Fig.~\ref{fig:travel_reward_vis} to explain why the second shot contributes disproportionately: refinement changes \emph{where} the generator places probability mass in synthetic space, not merely \emph{how much} it generates.

\paragraph{Round-0: diverse modes, but reward concentrated in a few easy regions.}
The round-0 generator already produces a multi-modal set of samples, visible as several separated clusters in the embedding space.
However, reward is highly uneven across these modes.
High-reward samples (yellow) concentrate in only a small subset of clusters, while many clusters remain uniformly low-reward (blue/gray).
Under extreme imbalance, this pattern is expected: without a utility-driven signal, synthetic sampling tends to overproduce majority-like cases that are schema-valid yet uninformative for minority-sensitive decisions.

\paragraph{Refinement induces two complementary effects: within-mode shifts and across-mode activation.}
Comparing round-0 and round-1, Fig.~\ref{fig:travel_reward_vis} suggests two distinct changes driven by reward-based refinement.
\emph{Within-mode shift:} in modes that persist across rounds, reward increases locally—blue/gray points are replaced by more gray/yellow points—indicating that refinement pushes samples toward reward-favorable subregions without collapsing the mode structure.
\emph{Across-mode activation:} reward is no longer confined to the same few modes; modes that were previously uniformly low-reward begin to contain high-reward samples.
This indicates that refinement does not merely oversample the easiest high-reward mode, but instead activates additional utility-relevant modes that were present but ineffective in round-0.

\paragraph{Why this explains the two-shot gain on the long tail.}
These two effects directly align with the observed performance shape on Travel.
The first shot adapts the predictor to the initial synthetic support, which is dominated by low-reward regions and therefore yields only a modest improvement.
The second shot benefits disproportionately because the generator has been reshaped to place more mass on reward-aligned regions \emph{within} modes and to expose the learner to a broader set of reward-bearing modes.
In a $p\approx0.01$ regime, this distributional change is precisely what enables minority-sensitive utility to improve: the learner sees more boundary-relevant and rare-event-supporting cases, rather than repeating majority-like patterns.

\paragraph{Implication for few-shot cross-silo coordination.}
The Travel analysis highlights why one-/two-shot coordination is meaningful in privacy-constrained deployments.
When the limiting factor is rare-event exposure rather than long-horizon optimization, a single refinement step that redistributes synthetic mass toward reward-aligned regions can translate into an immediate next-shot gain.
This mechanism is consistent with the design goal of \concordia: use reward feedback to reshape synthetic training under strict isolation, enabling reliable progress even when direct minority supervision is scarce.

\section{Conclusion and Broader Impact}

\paragraph{Conclusion.}
We introduced \textsc{Concordia}, a federated adaptation framework for tabular prediction when institutions cannot share raw records, validation sets, or generators, and must learn from client local synthetic tables.
\textsc{Concordia} closes a key gap in synthetic data enabled federated learning under heterogeneity because private imbalance aware validation utility becomes a training signal without being centralized.
Each client learns an on task utility scorer from local validation feedback to reweight synthetic batches for LoRA updates, while the server aggregates only task consistent LoRA deltas and redistributes scorer parameters as the utility proxy.
Across finance and healthcare, including an extreme long tail task Travel with p about 0.01 and evaluation on an unseen medical site VA Long Beach, \textsc{Concordia} improves MCC and reduces cross round volatility.
A generator audit shows refinement changes what is synthesized by preserving majority mode plausibility while strengthening coherent feature dependencies in finance, and sharpening multi feature risk conjunctions in healthcare, consistent with improved robustness under site shift.

\paragraph{Broader Impact.}
\textsc{Concordia} supports collaboration in regulated domains by keeping raw data, private validation sets, and generator parameters on premise, and may lower barriers for cross institution risk modeling and clinical decision support when labels are rare and institutions differ.
However, synthetic data and parameter efficient updates do not remove privacy risk.
Transmitted LoRA and scorer parameters may still leak information through membership inference or model inversion, and generators can memorize or amplify local biases.
We therefore view \textsc{Concordia} as an optimization framework rather than a complete privacy solution, and recommend secure aggregation, access control, and when appropriate clipping or noising of transmitted updates together with audits for subgroup fairness and bias amplification.
Finally, the same capability could be misused for harmful risk scoring against vulnerable groups, so responsible use requires governance, documentation of intended use, and continual monitoring in deployment.

%%
%% The next two lines define the bibliography style to be used, and
%% the bibliography file.

\bibliography{main}
\bibliographystyle{ACM-Reference-Format}

%%
%% If your work has an appendix, this is the place to put it.
\appendix
\section{Implementation Details}
\label{app:impl}

This section describes implementation details for reproducing \concordia, including the downstream model and LoRA configuration, client-local generators, scorers, GRPO refinement, and the federated protocol. Unless otherwise stated, all methods use the same backbone capacity and the same per-round compute and communication budget.

\subsection{Backbone LLM and Downstream Formulation}
\label{app:impl_backbone}

\textbf{Backbone.}
We use \textbf{Qwen3-4B} as the shared backbone for both the downstream tabular learner and the synthetic table generator.
The backbone is initialized from the \textbf{Qwen/Qwen3-4B} repository at revision \textbf{1cfa9a7208912126459214e8b04321603b3df60c}.

\textbf{Tabular task formulation.}
Each tabular record is serialized into a structured \textbf{JSON-like} representation (keys are double-quoted; values are numeric or unquoted categorical tokens) and paired with a task instruction that specifies \textbf{binary classification} and the label space.
The downstream model is trained using \textbf{causal LM loss over the label token} (i.e., the model directly generates a label string such as \texttt{healthy}/\texttt{unhealthy}).
During evaluation, the decision threshold is selected on the validation split to maximize MCC and then applied to the test split.

\subsection{Federated Protocol and Two-round Schedule}
\label{app:impl_protocol}

\textbf{Clients and rounds.}
We run $R=2$ communication rounds.
In healthcare, we simulate $N=3$ training clients (Cleveland, Hungary, Switzerland) under a single task identifier, with full participation in both rounds.
In finance, we run three independent single-client streams (German, Lending Club, Travel), each treated as its own task identifier; hence aggregation is not applicable within each finance stream.

At round $r$, the server broadcasts the task backbone $\theta_{\tau}^{(r)}$ to participating clients of task $\tau$.
Each participating client performs local LoRA adaptation on synthetic tables, updates its scorer using private validation feedback, refines its generator via GRPO, and uploads only its LoRA delta and scorer parameters.

\textbf{Task-aware aggregation.}
Clients are associated with a task identifier $\tau$, where \textbf{label semantics define task groups}.
The server aggregates LoRA deltas only within the same task identifier using \textbf{uniform average}.
Generator parameters are never uploaded or aggregated.
Scorers are collected and redistributed as a within-round pool and are not aggregated.

\subsection{LoRA Configuration and Local Optimization}
\label{app:impl_lora}

\textbf{LoRA modules.}
We apply LoRA to attention and MLP projections with
rank $r=\textbf{16}$, scaling $\alpha=\textbf{32}$, and dropout $\textbf{0}$, targeting
\texttt{[q\_proj, k\_proj, v\_proj, o\_proj, gate\_proj, up\_proj, down\_proj]}.
All backbone parameters remain frozen.

\textbf{Local steps and optimizer.}
Each client performs $T=\textbf{1500}$ local update steps per round (for a total of 3000 steps across two rounds).
We use \textbf{AdamW} with learning rate $\eta=\textbf{5e-5}$, weight decay $\textbf{0}$, and batch size $\textbf{4}$ synthetic samples per step.
We set the maximum sequence length to \textbf{1024} tokens and use gradient clipping with norm $\textbf{1.0}$.

\subsection{Client-local Synthetic Table Generator}
\label{app:impl_generator}

\textbf{Initialization.}
Each client maintains a client-local generator $G_{\theta_g^{(i)}}$, initialized from the same checkpoint family as the backbone, using the \textbf{instruct} variant (\textbf{Qwen3-4B-Instruct family}).
We do \textbf{not} perform public instruction tuning for schema adherence.

\textbf{Generation format.}
For each dataset, we construct a synthesis condition $c$ consisting of \textbf{(i) schema/field list, (ii) five in-class exemplar records, and (iii) a task instruction with explicit label strings}.
The generator outputs a single serialized record in \textbf{JSON-like} format.
We validate outputs using the \textbf{FormatReward} parser/reward (brace-block extraction + required key/label checks) and discard malformed samples.

\textbf{Sampling parameters.}
We generate synthetic tables using temperature $\textbf{1.0}$, top-$p$ $\textbf{0.95}$, and maximum new tokens $\textbf{512}$.
Each local update step samples \textbf{4} synthetic tables (matching the LoRA batch size).

\subsection{Scorer Model and Utility-weighted Training}
\label{app:impl_scorer}

\textbf{Scorer architecture.}
Each client maintains a lightweight scorer $S_{\phi_i}$ that maps a serialized synthetic table to a scalar score.
We implement the scorer by attaching a \textbf{scorer LoRA adapter} to the backbone and predicting a scalar from the final hidden representation; the scorer adapter uses
rank $r=\textbf{8}$, scaling $\alpha=\textbf{16}$, dropout $\textbf{0.05}$, targeting \texttt{[q\_proj, v\_proj]}.

\textbf{Input and representation.}
The scorer consumes \textbf{only the JSON-like string}.
We reuse backbone representations by extracting the \textbf{last layer} hidden state (last-token pooling) as the scorer input.

\textbf{Weighting function.}
The scorer outputs a bounded value $S_{\phi_i}(x)\in[-1,1]$ via $\tanh$.
For a mini-batch $X_{\mathrm{syn}}$, we map scores to nonnegative weights by
\[
w_{\phi_i}(X_{\mathrm{syn}})=
\mathrm{Norm}\!\left(
\left\{
\mathrm{clip}\!\left(\frac{S_{\phi_i}(x)+1}{2},\, w_{\min},\, w_{\max}\right)
\right\}_{x\in X_{\mathrm{syn}}}
\right),
\]
with $(w_{\min},w_{\max})=(0.05,0.95)$ and $\mathrm{Norm}$ normalizing weights to sum to one within the mini-batch.

\textbf{Scorer update frequency.}
We refresh scorer parameters every $m=\textbf{15}$ local LoRA steps.
The validation utility is computed on the private validation set using $\mathcal{L}_{\mathrm{val}} = 1-\mathrm{MCC}$.
We optimize the scorer with \textbf{AdamW} and learning rate $\eta_\phi=\textbf{5e-5}$.

\subsection{Within-round Scorer Pool and Cross-client Evaluation}
\label{app:impl_pool}

At the end of each round, participating clients upload scorer parameters to the server.
The server constructs a within-round scorer pool
\[
\mathcal{S}_r = \{S_{\phi_j}\}_{j \in \mathcal{C}_r},
\]
and redistributes $\mathcal{S}_r$ to all clients in $\mathcal{C}_r$.
Scorers are used only for evaluation and are never fine-tuned on other clients' data.

To combine heterogeneous scorer outputs, we use \textbf{mean pooling} over group-normalized scores.
Specifically, for a candidate group of size $G$, each scorer contributes only a within-group relative preference signal using \textbf{rank normalization}.

\subsection{Generator Refinement with GRPO}
\label{app:impl_grpo}

\textbf{Candidate groups.}
For each synthesis condition $c$, the client samples a group of $G=\textbf{8}$ candidate tables.
We evaluate each candidate with the scorer pool and compute an ensemble reward using \textbf{mean} over normalized scores.

\textbf{Advantages and stabilization.}
We compute group-relative advantages by subtracting the within-group mean.
We normalize advantages within each group and clip to $\textbf{5.0}$.

\textbf{GRPO objective details.}
We use a frozen reference policy equal to the generator snapshot at the beginning of each round and apply a KL penalty to stabilize updates.
The GRPO update is sample-level: each completion receives a group-relative advantage computed from the composed reward, normalized within the group.
We use a PPO-style clipped objective with clip range 0.2 and KL coefficient 0.01 (kept fixed across tasks).

\textbf{GRPO updates.}
We update the client-local generator using GRPO with learning rate $\eta_g=\textbf{1e-6}$.
Per round, we run \textbf{300} GRPO update steps over \textbf{600} synthesis conditions (batch size 2 prompts per step), using maximum prompt length \textbf{1024}, maximum completion length \textbf{512}, temperature \textbf{1.0}, and top-$p$ \textbf{0.95}.
We use stop strings \{\texttt{\}}\texttt{\textbackslash n}, \texttt{\}}\texttt{\textbackslash r\textbackslash n}, \texttt{\}}\texttt{\textbackslash n\textbackslash n}, \texttt{\}}\} and batch reward scaling.
The generator is never uploaded.

\subsection{Random Seeds and Reporting}
\label{app:impl_reporting}

We report MCC as mean $\pm$ standard deviation over \textbf{3} random seeds (seed $\in\{0,1,2\}$).
Each seed controls synthetic prompt construction and sampling, model/adapter initialization, data shuffling, and GRPO sampling.
For each seed, we run exactly two communication rounds and evaluate after each round.
We use the same number of synthetic samples and the same local optimization steps for all methods and ablations.

\section{Prompts and Tabular Serialization}
\label{app:prompts}

This section documents the tabular serialization format and the client-local prompting templates used for synthetic table generation.

\subsection{Serialization Format}
\label{app:serialization}

Each record is serialized into a \textbf{JSON-like} format with a single record per sample.
We use a fixed key order to avoid format drift across rounds.
Missing values are encoded as \textbf{\texttt{nan}}.
Categorical values are represented as \textbf{unquoted tokens} (e.g., \texttt{good}, \texttt{bad}) and continuous values are represented with \textbf{4 decimal places}.

\paragraph{Task-specific instructions (verbatim).}

\begin{promptcard}{Heart disease}
\ttfamily\small
Predict whether the patient has heart disease using the features below: Age, Sex, ChestPainType, RestingBloodPressure, Cholesterol, FastingBloodSugar, RestingECG, MaxHeartRate, ExerciseInducedAngina, OldPeak, PeakExerciseSTSegmentSlope, NumberOfVesselsColored, and ThalassemiaStatus. Directly respond with 'unhealthy' if the patient has heart disease or 'healthy' if they do not.
\end{promptcard}

\begin{promptcard}{Credit risk}
\ttfamily\small
Predict whether the loan applicant is a credit risk using the features below. Directly respond with 'bad' if the applicant is a bad credit risk or 'good' if they are a good credit risk.
\end{promptcard}

\begin{promptcard}{Loan status}
\ttfamily\small
Predict the loan status using the features below. Directly respond with 'fullypaid' if the loan is fully repaid or 'chargeoff' if the loan is charged off.
\end{promptcard}

\begin{promptcard}{Insurance claim}
\ttfamily\small
Predict whether the customer will file an insurance claim based on the features below. Directly respond with 'yes' if the customer is likely to make a claim, or 'no' if the customer is unlikely to make a claim.
\end{promptcard}

\subsection{Synthesis Condition}
\label{app:synthesis_condition}

For each client, the synthesis condition includes the schema (field list), in-class exemplar records, and the task specification.
The task specification defines the prediction target and the label format, e.g.,
\texttt{healthy/unhealthy} (heart disease), \texttt{good/bad} (credit risk), \texttt{fullypaid/chargeoff} (loan status), and \texttt{yes/no} (insurance claim).

\subsection{Generator Prompt Template}
\label{app:prompt_template}

We use the following client-local prompt template (5-shot in-class exemplars):

\begin{quote}
\textbf{System:} You are a data synthesis assistant. Output only one valid JSON-like object that matches the schema. \\
\textbf{User:} Here are five entries of tabular data in JSON format, each consisting of \emph{$n$} features. Each feature is described as ``\texttt{"feature name": value}''. The target feature \emph{label} is a classification task.\\
\textbf{User:} \texttt{sample one:} \ldots (5 exemplars) \\
\textbf{User:} Directly generate only one new final sample in JSON format that approximates the key patterns observed in the provided samples.\\
\textbf{Assistant:} \texttt{\{ ... \}}
\end{quote}

We apply a strict parser/reward (FormatReward) to the output and discard malformed generations.

\section{Communication and Compute Cost}
\label{app:cost}

\subsection{Communication Payload}
\label{app:comm}

Each client uploads (i) the downstream LoRA delta $\Delta\theta_i$ and (ii) the scorer parameters $\phi_i$.
We communicate only a \textbf{scorer LoRA adapter + scalar scorer head}; the scorer backbone is the shared public LLM checkpoint (no task head, no full scorer weights; generators are never transmitted).
Using bf16 transmission (2 bytes/parameter), for Qwen3-4B ($L{=}36$, $d{=}2560$):
\[
\begin{aligned}
P_{\mathrm{lora}} &= \textbf{33{,}030{,}144},\\
P_{\mathrm{s}} &= 4Lr_s d = \textbf{2{,}949{,}120},\\
P_{\mathrm{head}} &= d+1 = \textbf{2{,}561}.
\end{aligned}
\]

\paragraph{Uplink (client $\rightarrow$ server).}
Per-client uplink:
\[
2\,(P_{\mathrm{lora}}+P_{\mathrm{s}}+P_{\mathrm{head}})
=\textbf{71.96 MB}\ (\textbf{68.64 MiB}),
\]
so with $|\mathcal{C}_r|{=}3$ clients the total uplink per round is
$\textbf{215.89 MB}$ ($\textbf{205.93 MiB}$).

\paragraph{Downlink (server $\rightarrow$ clients).}
Per-client backbone broadcast (aggregated LoRA delta):
\[
2P_{\mathrm{lora}}=\textbf{66.06 MB}\ (\textbf{62.99 MiB}).
\]
If the within-round scorer pool is redistributed, each client receives the other $|\mathcal{C}_r|-1$ scorers:
\[
2(|\mathcal{C}_r|-1)(P_{\mathrm{s}}+P_{\mathrm{head}})
=\textbf{11.81 MB}\ (\textbf{11.26 MiB})\ \text{for }|\mathcal{C}_r|{=}3.
\]
We exclude one-time distribution of the shared backbone checkpoint.

\section{Privacy Discussion and Threat Model}
\label{app:privacy}

\subsection{What Remains Local}
\label{app:privacy_local}

Raw training records, private validation sets, and generated synthetic tables remain on the client.
Generator parameters remain on the client and are never uploaded.
Communication is restricted to (i) downstream LoRA deltas for the task backbone and (ii) scorer parameters.

\subsection{Scorer Sharing and Privacy Boundary}
\label{app:privacy_scorer}

Scorers are shared as evaluators to enable validation-aligned coordination without centralizing any validation data.
Crucially, a communicated scorer does \textbf{not} include a separate task head or the full scorer LLM weights:
all parties share the same public backbone checkpoint, and each client only transmits a \textbf{scorer LoRA adapter}
plus a \textbf{scalar scorer head}.
This significantly reduces the communicated surface compared to sharing full evaluator models, while still enabling
cross-client preference signals for generator refinement.

That said, the scorer remains a learned model component and could in principle leak information about the client
distribution under strong adversarial assumptions.
We therefore treat scorer sharing as a pragmatic privacy tradeoff:
it avoids transmitting raw validation examples and generator parameters, but introduces a limited model-sharing surface.
% \section{Research Methods}

% \subsection{Part One}

% Lorem ipsum dolor sit amet, consectetur adipiscing elit. Morbi
% malesuada, quam in pulvinar varius, metus nunc fermentum urna, id
% sollicitudin purus odio sit amet enim. Aliquam ullamcorper eu ipsum
% vel mollis. Curabitur quis dictum nisl. Phasellus vel semper risus, et
% lacinia dolor. Integer ultricies commodo sem nec semper.

% \subsection{Part Two}

% Etiam commodo feugiat nisl pulvinar pellentesque. Etiam auctor sodales
% ligula, non varius nibh pulvinar semper. Suspendisse nec lectus non
% ipsum convallis congue hendrerit vitae sapien. Donec at laoreet
% eros. Vivamus non purus placerat, scelerisque diam eu, cursus
% ante. Etiam aliquam tortor auctor efficitur mattis.

% \section{Online Resources}

% Nam id fermentum dui. Suspendisse sagittis tortor a nulla mollis, in
% pulvinar ex pretium. Sed interdum orci quis metus euismod, et sagittis
% enim maximus. Vestibulum gravida massa ut felis suscipit
% congue. Quisque mattis elit a risus ultrices commodo venenatis eget
% dui. Etiam sagittis eleifend elementum.

% Nam interdum magna at lectus dignissim, ac dignissim lorem
% rhoncus. Maecenas eu arcu ac neque placerat aliquam. Nunc pulvinar
% massa et mattis lacinia.

\end{document}